\documentclass[conference]{IEEEtran}
\IEEEoverridecommandlockouts
\usepackage{cite}
\usepackage{amsmath,amssymb,amsfonts}
\usepackage{amsthm}
\usepackage{algorithmic}
\usepackage{graphicx}
\usepackage{textcomp}
\usepackage{xcolor}
\usepackage{mathrsfs}
\renewcommand{\Re}{\operatorname{Re}}
\renewcommand{\Im}{\operatorname{Im}}
\def\BibTeX{{\rm B\kern-.05em{\sc i\kern-.025em b}\kern-.08em
    T\kern-.1667em\lower.7ex\hbox{E}\kern-.125emX}}
\begin{document}

\title{Mobile Robot Navigation in Complex Polygonal Workspaces Using Conformal Navigation Transformations}

\author{\IEEEauthorblockN{Li Fan}
\IEEEauthorblockA{\textit{College of Information Science and Engineering} \\
\textit{Northeastern University}\\
Shenyang 110819, China \\
lifanneu@yeah.net}
\and
\IEEEauthorblockN{Jianchang Liu\textsuperscript{*}}
\IEEEauthorblockA{\textit{College of Information Science and Engineering} \\
\textit{Northeastern University}\\
Shenyang 110819, China \\
liujianchang@ise.neu.edu.cn}
}

\maketitle

\begin{abstract}

%Navigation functions provide both path and motion planning, which can be used to ensure obstacle avoidance and convergence in the sphere world. When dealing with complex and realistic scenarios, constructing a transformation to the sphere world is essential and, at the same time, challenging. %When dealing with complex and realistic scenarios, constructing a transformation to the sphere world is both essential and challenging.
%Navigation functions provide both path and motion planning, which can be used to achieve collision-free navigation of mobile robots.
%The navigation function method can simultaneously solve the path and motion planning problems of mobile robots. When dealing with complex and realistic scenarios, it mostly relies on a diffeomorphism transformation to the sphere world.
%The navigation function method can simultaneously solve the path and motion planning problems of mobile robots, and it mostly relies on a diffeomorphism transformation to the sphere world when handling complex and realistic scenarios.
This work proposes a novel transformation termed the conformal navigation transformation to achieve collision-free navigation of a robot in a workspace populated with arbitrary polygonal obstacles. The properties of the conformal navigation transformation in the polygonal workspace are investigated in this work as well as its capability to provide a solution to the navigation problem. %The properties of the conformal navigation transformation are investigated, which contribute to the solution of the robot navigation problem in complex polygonal environments. %which facilitates the navigation of robots in complex environments. 
The definition of the navigation function is generalized to accommodate non-smooth obstacle boundaries. Based on the proposed transformation and the generalized navigation function, a provably correct feedback controller is derived for the automatic guidance and motion control of the kinematic mobile robot. Moreover, an iterative method is proposed to construct the conformal navigation transformation in a multi-connected polygonal workspace, which transforms the multi-connected problem into multiple single-connected problems to achieve fast convergence. % an iterative construction method
In addition to the analytic guarantees, the simulation study verifies the effectiveness of the proposed methodology in a workspace with non-trivial polygonal obstacles.
%The theoretical results provided are supported by analytical proofs. Simulation studies are conducted to demonstrate the effectiveness of the methodology.%control scheme

%The navigation function method can simultaneously solve the path and motion planning problems of mobile robots, and it mostly relies on a diffeomorphism transformation to the sphere world when handling complex and realistic scenarios.This work proposes a novel transformation termed the conformal navigation transformation to achieve collision-free navigation of a robot in a workspace populated with obstacles of arbitrary shapes. The properties of the conformal navigation transformation, including uniqueness, invariance of navigation properties and no angular deformation, are investigated in this work, which facilitates the navigation of robots in complex environments. Based on navigation functions and conformal navigation transformations, feedback controllers are derived for automatic guidance and motion control of kinematic and dynamic mobile robots. Moreover, an iterative construction method for the conformal navigation transformation on a multi-connected workspace is proposed, which transforms the multi-connected problem into multiple single-connected problems to achieve fast convergence. The theoretical results provided are supported by analytical proofs. Simulation studies are conducted to demonstrate the effectiveness of the proposed control scheme.
\end{abstract}

%\begin{IEEEkeywords}
%Motion and path planning, collision avoidance, navigation function, polygonal obstacle, conformal map
%\end{IEEEkeywordskeywords}

%%%%%%%%%%%%%%%%%%%%%%%%%%%%%%%%%%%%%%%%%%%%%%%%%%%%%%%%%%%%%%%%%%%%%%%%%%%%%%%%
\section{INTRODUCTION}

The autonomous navigation of robots in cluttered environments is an actively studied topic of robotics research. Control laws based on artificial potential fields (APFs) constitute a widely researched tool for solving the robot navigation problem due to their intuitive design and ability to simultaneously solve path planning and motion planning subproblems \cite{1,2}. On the other hand, this control method suffers from the presence of local minima and the inability to cope well with non-convex obstacles.%this control method suffers from the problem of local minima, and cannot cope well with non-convex obstacles.

The navigation function \cite{Rimon1992,6,7} method proposed by Rimon and Koditschek is an efficient method for constructing a class of local minimal free APFs in the sphere world, which can be extended to environments with star-shaped obstacles by constructing a diffeomorphism. The negative gradient of the navigation function can be served as a control input to guarantee collision-free motion of the kinematic mobile robot and convergence to the goal point from almost all initial points. A large number of successful applications have appeared in the literature, such as multi-robot systems \cite{Loizou2008,11,12}, highly dynamic or partially known environments \cite{Li2019,Loizou2022}, uncertain dynamic systems \cite{15}, and constrained stabilization problems \cite{16}. In a recent work \cite{17}, the construction of navigation functions was extended to non-spherical convex obstacles; however, the construction of navigation functions in geometrically complex spaces is still a challenging problem. %however constructing

Transforming geometrically complex spaces to geometrically simple spaces is an important and elegant approach to implementing navigation functions in complex workspaces. Techniques based on diffeomorphism and navigation transformation are a few of the promising methods found in the literature to achieve the above objective. In \cite{6}, a family of analytic diffeomorphisms was constructed for mapping any star world %$\footnote{A star set is a set that possesses a point from which all the rays cross the boundary only once. A star world is formed by removing from the interior of a large star set a finite number of non-overlapping smaller star sets.}$ 
to an appropriate sphere world, requiring appropriate tuning of a certain parameter. To cope with the complexity of the workspace, the authors in \cite{18} proposed a diffeomorphism based on the harmonic map, which maps the workspace to a punctured disc by solving multiple boundary problems. By applying a two-step navigation transformation, the author in \cite{19} achieved a tuning-free solution by pulling back a trivial solution in the point world to the initial star world. In a continuous work \cite{Nicolas2020}, the authors propose a single-step navigation transformation candidate that directly maps a known star world to a point world.

The transformation candidates in most of the aforementioned works are only suitable for smooth star worlds rather than non-smooth workspaces, which hinders the application of navigation functions in realistic environments. %Most of the aforementioned works depend on a transformation from the workspace to the sphere world, but the transformation candidates are mainly suitable for the star world rather than an arbitrary workspace. %applicable to
In this work, we address the problem of navigating a kinematic mobile robot system in a workspace with arbitrary polygonal obstacles and connectedness by constructing a novel transformation termed the conformal navigation transformation. Unlike the transformations mentioned above, the proposed transformation requires neither tuning a parameter nor decomposing the workspace into trees of stars, and it is unique in a given workspace. Furthermore, the class of navigation functions defined in smooth workspaces is extended to accommodate polygonal obstacles. Then, the proposed transformation can pull back the extended navigation function in the sphere world to the initial workspace, as well as the trivial solution in the sphere world determined by the vector field, so that the navigation problem in the polygonal workspace can be solved elegantly. Finally, we present a fast iterative method for constructing the proposed transformations in any complex workspace. To the best of the authors' knowledge, this is the first work to provide a tuning-free transformation candidate that maps any workspace with non-star-shaped polygonal obstacles to a sphere world. 
%In addition, we present a method to obtain conformal navigation transformations in multi-connected workspaces by iterating over the conformal navigation transformations in single-connected workspaces, while also reducing the computational load required to deal with complex workspaces.

The rest of this paper is organized as follows. Section \uppercase\expandafter{\romannumeral2} introduces the necessary preliminaries. Section \uppercase\expandafter{\romannumeral3} analyzes the properties of the conformal navigation transformation. Section \uppercase\expandafter{\romannumeral4} presents a mobile robot controller designed using the proposed transformation and the navigation function. Section \uppercase\expandafter{\romannumeral5} proposes the construction method of the conformal navigation transformation. Section \uppercase\expandafter{\romannumeral6} provides the simulation study to demonstrate the effectiveness of the proposed control scheme. Section \uppercase\expandafter{\romannumeral7} concludes this paper.

\section{Preliminaries}
This section gives the necessary terminology and definitions for the development of the methodology.
%In this section, the necessary terminology and definitions for the development of the methodology are introduced.

\subsection{Sphere Worlds and Workspaces}
The $n$-dimensional sphere world $\mathscr{M}^n$ as defined in \cite{Rimon1992} is a compact connected subset of ${\mathbb{R}^n}$ whose boundary is formed by the disjoint union of an external $(n - 1)$-sphere and $M\in{\mathbb{N}}$ internal $(n - 1)$-spheres. Each internal sphere obstacle ${\mathscr{\tilde{O}}_i}$ is the interior of the $(n - 1)$-sphere, which is implicitly defined as $\mathscr{\tilde{O}}_i=\{q\in\mathbb{R}^n:\|q-q_i\|^2<\rho_i^2\},\ i\in\{1,...,M\}.$ For simplicity, the exterior of the external $(n - 1)$-sphere is referred to as the zeroth sphere obstacle, denoted as ${\mathscr{\tilde{O}}_0}$. In this regard, the $n$-dimensional sphere world with $M$ internal obstacles $\mathscr{M}^n$ is represented as:%\begin{equation}\begin{split}\label{eq1}\mathscr{M}_M^n=\{q\in\mathbb{R}^n:&\|q-q_0\|^2\leqslant\rho_0^2,\\&\|q-q_1\|^2\geqslant\rho_1^2,...,\|q-q_M\|^2\geqslant\rho_M^2\}.\end{split}\end{equation}%
\begin{equation}\begin{split}\label{eq1}\mathscr{M}^n=\{&q\in\mathbb{R}^n:-\|q-q_0\|^2\geqslant\rho_0^2,\\&\|q-q_1\|^2\geqslant\rho_1^2,...,\|q-q_M\|^2\geqslant\rho_M^2\}.\end{split}\end{equation}

Clearly, the sphere world is geometrically simple and can be used as a topological model of geometrically complex spaces. Formally, the geometrically complex robot workspace is defined as follows:

\textit{Definition 1:} The $n$-dimensional robot workspace ${\mathscr{W}^n}$ is a connected and compact $n$-dimensional manifold with boundary such that ${\mathscr{W}^n}$ is homeomorphic to $\mathscr{M}^n$.

For a finite $M\in{\mathbb{N}}$, the internal obstacle ${\mathscr{O}_i}$ of ${\mathscr{W}^n}$ is the interior of a connected and compact $n$-dimensional subset of ${\mathbb{R}^n}$ such that $\mathscr{O}_i\cap\mathscr{O}_j=\varnothing,\ i\ne j,\ i,j\in\{1,...,M\}.$ It is convenient to regard the unbounded component of the workspace's complement as the zeroth obstacle. Each obstacle can be represented by an obstacle function $\beta_i:{\mathbb{R}}^n\to\mathbb{R}$ in the following form: 
\begin{equation}\label{workspace obstacle}\mathscr{O}_i=\{q\in\mathbb{R}^n:\beta_i(q)<0\},\ i\in\{0,...,M\}.\end{equation}

For the obstacle function, zero is not a critical value and the zero level set of $\beta_i$ defines the obstacle's boundary: $\partial\mathscr{O}_i=\{q\in\mathbb{R}^n:\beta_i(q)=0\},\ i\in\{0,...,M\}$. Moreover, according to the implicit function theorem, the boundary of an obstacle is an $(n-1)$-dimensional  submanifold of $\mathscr{W}^n$. 
Then, the $n$-dimensional workspace with $M$ internal obstacles ${\mathscr{W}^n}$ is represented as: \begin{equation}\label{workspace}\mathscr{W}^n=\{q\in\mathbb{R}^n:\beta_0(q)\geqslant0,...,\beta_M(q)\geqslant0\}.\end{equation}

\subsection{Polyhedral Obstacles and Polyhedral Workspaces}

In realistic environments, obstacles may have ``vertices" even in the simplest scenarios. Here, we will restrict our attention to the polyhedral obstacle and give a formal definition.
%We will restrict our attention to the geometrically complex polyhedral obstacle and the polyhedral workspace, which allow the obstacles to be non-star-shaped polyhedrons.

Let $H_i$ be a open half-space in $\mathbb{R}^n$, $q_i$ be any point on $H_i$, and $n_i$ be the unit normal vector of $H_i$. The direction of $n_i$ is assumed to be outward pointing with respect to the polyhedron. Each half-space can be represented by a linear inequality in the following form:\begin{equation}\label{obstacle function for half-space}H_i=\{q\in\mathbb{R}^n:(q-q_i)\cdot{n_i}<0\}.\end{equation}
A set $\mathscr{O}_i\subset\mathbb{R}^n$ is polyhedral if it can be expressed as a finite Boolean combination (via the set operations  $\cup, \cap, -, c$) of half-spaces. Since set union and difference can be reduced to intersection and complement, polyhedrons can be generated from a finite number of half-spaces by intersection and complement operations. %Set union and difference can be reduced to intersection and complement. Thus, polyhedrons can be generated from a finite number of half-spaces by set intersection and set complement operations. 
In particular, convex polyhedrons can be defined as the intersection of a finite number of half-spaces. Formally, the polyhedral obstacle is defined as follows:

\textit{Definition 2:} A polyhedral obstacle, $\mathscr{O}_i$, is a polyhedral set in $\mathbb{R}^n$ comprised of a finite intersection and complement of half-spaces, $H_{ij}$ for $j\in\{1,...,M\}$, such that:
\begin{enumerate}
\item $\mathscr{O}_i$ is the interior of a connected and compact $n$-dimensional manifold with boundary;
\item the intersection of the boundaries of any two intersecting half-space, $\partial H_{ij}\cap\partial H_{ik}$, is an $(n-2)$-dimensional subset of $\mathbb{R}^n$.
%the boundaries of any two intersecting polyhedral sets, $\partial H_{ij}$ and $\partial H_{ik}$, meet transversally.
\end{enumerate}
If all the obstacles of the workspace are polyhedral obstacles, then the workspace is called a polyhedral workspace, denoted by $\mathscr{P}^n$. In particular, we refer to the polyhedral obstacles and polyhedral workspaces in $\mathbb{R}^2$ as polygonal obstacles and polygonal workspaces, respectively.

The intersection of $\partial H_{ij}$ and $\partial\mathscr{O}_i$ is an $(n-1)$-dimensional subset of $\mathbb{R}^n$, called the facet of the polyhedral obstacle, denoted by $\mathscr{F}_{ij}$. The boundary of all the facets in $\partial\mathscr{O}_i$, $$\mathscr{V}_i=\partial\mathscr{O}_i-\bigcup_{j=1}^M\mathring{\mathscr{F}}_{ij},$$ is the set of vertices. The obstacle function $\beta_i$ for the polyhedral obstacle is an analytical function away from the vertices of $\mathscr{O}_i$. In general, $\beta_i$ for the polyhedral obstacle can be constructed automatically based on the function \eqref{obstacle function for half-space} for each half-space and the Boolean combination used to construct $\mathscr{O}_i$. 

\subsection{Navigation Function}
%The navigation function is a well-established technique for robot navigation in spaces populated with obstacles. For almost all initial states, the control law generated by the navigation function defines asymptotically stable closed-loop robot systems with collision-free trajectories.
The navigation function is a well-established technique for robot navigation in a smooth manifold with boundary \cite{7}. However, the polyhedral workspace is not smooth because its tangent space at the vertices of the polyhedral obstacles is not well defined. In this case, the navigation function cannot be smooth over the whole workspace. This paper defines a non-smooth version of the navigation function that relaxes the smooth property.

\textit{Definition 3 :} Let $\mathscr{P}^n\subset\mathbb{R}^n$ be a $n$-dimensional polyhedral workspace, $\mathscr{V}\subset\partial\mathscr{P}^n$ be the set of vertices of the polyhedral obstacles, and $x_d$ be the goal point in the interior of $\mathscr{P}^n$. A map $\phi:\mathscr{P}^n\to[0,1]$, is a navigation function if it:
\begin{enumerate}
	\item is continuous on $\mathscr{P}^n$ and analytic on $\mathscr{P}^n-\mathscr{V}$.
	\item is admissible on $\mathscr{P}^n$: uniformly maximal on $\partial\mathscr{P}^n$.
	\item is polar at $x_d$: has a unique minimum at $x_d$.
	\item is morse on $\mathscr{P}^n-\mathscr{V}$: has only non-degenerate critical points on $\mathscr{P}^n-\mathscr{V}$.
	\item has a bounded gradient on $\mathscr{P}^n-\mathscr{V}$.
\end{enumerate}

The Koditschek-Rimon navigation function $\phi_{kr}=\frac{\|q-q_d\|}{{(\|q-q_d\|^k+\beta)}^{1/k}}$, is a smooth navigation function constructed in the sphere world, so long as $k$ exceeds some threshold, where $\beta=\prod_{j=0}^M\beta_i$ is the aggregate obstacle function \cite{7}.

\subsection{Mobile Robot Model}
This work considers a mobile robot that navigates in a $2$-dimensional polygonal workspace. The motion of the mobile robot is described by the trivial kinematic integrator model:
\begin{equation}\dot{x}(t)=u(t),\label{kinematic}\end{equation}
where $x\in\mathbb{R}^2$ is the robot position and $u\in\mathbb{R}^2$ is the corresponding control input vector. The initial position and goal position of the mobile robot are denoted as $x_0\in\mathscr{P}^2$ and $x_d\in\mathscr{P}^2$, respectively.

\subsection{Conformal Navigation Transformation}

Let $U$ be a connected open set, and $z=x+iy$ be a complex variable. A complex-valued function $f(z):U\subset\mathbb{C}\to\mathbb{C}$ is holomorphic on $U$ if it is complex differentiable at every point of $U$. Let $f(x,y)=(u(x,y),v(x,y)):U\subset\mathbb{R}^2\to\mathbb{R}^2$ be of class $C^1$ with non-vanishing Jacobian, then $f$ is a conformal map if and only if $f(z)=u(x,y)+iv(x,y):U\subset\mathbb{C}\to\mathbb{C}$, as a function of $z$, is holomorphic \cite{20}.

The conformal navigation transformation defined below provides an efficient way to implement the navigation function in a geometrically complex polygonal workspace.

\textit{Definition 4:} A conformal navigation transformation is a map $T:\mathscr{W}^2\to\mathscr{M}^2$ which:
\begin{enumerate}
	\item maps the interior of the workspace conformally to the interior of a $2$-dimensional sphere world.
	\item maps the external boundary of the workspace homeomorphically to the unit circle.
	\item maps the boundaries of $M$ internal obstacles homeomorphically to $M$ small disjoint circles inside the unit circle.
\end{enumerate}

\section{Conformal Navigation Transformation Analysis}

This section investigates the properties of the conformal navigation transformation on the polygonal workspace. These properties enable the proposed transformation to be applied to the navigation problem.

%This section investigates properties of the conformal navigation transformation that enable its application to robot navigation problems in arbitrary polygonal workspaces.

The following proposition establishes that the proposed transformation exists in any 2-dimensional workspace and provides the basis for constructing it on the polygonal space.

\textit{Proposition 1 \cite{FanLi2022}:} If $\mathscr{W}^2$ is a $2$-dimensional workspace with $M$ internal obstacles, then there exists a $2$-dimensional sphere world $\mathscr{M}^2$ with $M$ internal obstacles and a conformal navigation transformation $T$, such that $T(\mathscr{W}^2)=\mathscr{M}^2$.

The next proposition describes the uniqueness of the proposed transformation corresponding to a given workspace.
%The next proposition shows that there exists a unique conformal navigation transformation corresponding to a given workspace.

\textit{Proposition 2 \cite{FanLi2022}:} The conformal navigation transformation $T:\mathscr{W}^2\to\mathscr{M}^2$ and $\mathscr{M}^2$ are uniquely determined by ${\mathscr{W}^2}$ if the images of  the prescribed three points on the external boundary $\partial\mathscr{O}_0$ are fixed  on the unit circle.

The uniqueness of the proposed transformation implies that no further parameter tuning is required after the construction of the conformal navigation transformation; that is, the resulting solution is correct-by-construction. 

The above results are applicable to %are valid for% hold for%
all 2-dimensional workspaces $\mathscr{W}^2$. However, the vertices of polygonal obstacles introduce a new technical difficulty: the boundaries of polygonal obstacles are not differential manifolds. To address the problem caused by the vertices, we continue to investigate the properties of $T$ on the polygonal workspace $\mathscr{P}^2$.

%The invariance of the navigation properties in [1] states that the navigation function $\phi$ in a sphere world can be pulled back to a smooth workspace by the proposed transformation. %The navigation function in a sphere world can be pulled back to a smooth workspace by the proposed transformation, as stated in 1. 
%The fact that polygonal obstacles are not even differentiable manifolds due to the existence of vertices requires a reconsideration of the above property. The following results restated the invariance of the navigation properties in detail.

The machinery of complex variables provides a convenient way to study $T$ on $\mathscr{P}^2$, which will be used in the following. In fact, $T(x)$ on $\mathscr{P}^2$ and $T(z)$ on the complex plane satisfy the relation: $T(x)=[\Re(T(z)),\Im(T(z))]$. The next proposition establishes the analyticity of $T$ on $\mathscr{P}^2-\mathscr{V}$.

\textit{Lemma 1:} The conformal navigation transformation $T:\mathscr{P}^2\to\mathscr{M}^2$ is analytic on $\mathscr{P}^2-\mathscr{V}$.

\textit{Proof:} For the workspace with multiple obstacles, $T$ has the same boundary behavior as the workspace without internal obstacles \cite{pommerenke1992}. Without loss of generality, let polygon $\partial\mathscr{O}_0$ be the only boundary of the ${\mathscr{P}^2}$ without internal obstacles and $\mathscr{V}_0$ be the set of vertices of $\mathscr{O}_0$. Since the rotation does not change the analyticity of $T$, we may as well assume that one of the edges of $\partial\mathscr{O}_0$ lies on the real axis, denoted as $\partial\mathscr{O}_0^1$. Notice that when $x\in \partial\mathscr{O}_0^1$ is a non-vertex point, there is a disk around $x$ such that the half disk in ${\mathscr{P}^2}$ does not contain the point $z_0$ with $T(z_o) = 0$. Then $\log T(z)$ has a single-valued branch when $z$ is in the half disk. Since $|T(z)|$ tends to 1 as $z$ approaches $\partial\mathscr{O}_0^1$, the real part of $\log T(z)$ tends to 0. By the Reflection Principle \cite{pommerenke1992}, $\log T(z)$ has an analytic extension to the whole disk. Taking exponential, we conclude that $T$ is an analytic function at $x\in \partial\mathscr{O}_0^1-\mathscr{V}_0$. Rotating all other edges of $\partial\mathscr{O}_0$ to the real axis in turn, we have that $T$ is analytic on $\partial\mathscr{O}_0-\mathscr{V}_0$. Since the analyticity of $T$ on $\partial\mathscr{P}^2-\mathscr{V}$ is the same as that on $\partial\mathscr{O}_0-\mathscr{V}_0$, $T$ is analytic on $\partial\mathscr{P}^2-\mathscr{V}$. Hence, $T$ is analytic on $\mathscr{P}^2-\mathscr{V}$.$\hfill\qedsymbol$

Lemma 1 implies that $\phi\circ T$ is analytic on $\mathscr{P}^2-\mathscr{V}$. The next lemma and propositions show that the Jacobian of $T$ is non-singular on $\mathscr{P}^2-\mathscr{V}$.

\textit{Lemma 2:}  The conformal navigation transformation $T:\mathscr{P}^2\to\mathscr{M}^2$ is injective on $\mathscr{P}^2$.

\textit{Proof:} By definition, $T(z)$ is injective on $\partial\mathscr{P}^2$. Since $T(z)$ is a conformal equivalence from $\mathring{\mathscr{P}^2}$ onto $\mathring{\mathscr{M}^2}$, $T(z)$ is injective on $\mathring{\mathscr{P}^2}$\cite{FanLi2022}. Hence, $T$ is injective on $\mathscr{P}^2$.$\hfill\qedsymbol$

Based on the injectivity of $T$, the following proposition establishes that $T$ has a diffeomorphic extension to boundary points other than vertices.

\textit{Proposition 3:} The conformal navigation transformation $T:\mathscr{P}^2\to\mathscr{M}^2$ is an analytic diffeomorphism on $\mathscr{P}^2-\mathscr{V}$.

\textit{Proof:} Since an injective map cannot be constant, then according to the open mapping theorem, $T$ is an open map. Hence, the map $T^{-1}$ is continuous. From the Inverse function theorem, we know that $T^{-1}$ is analytic in the set of all $w=T(z)$ with $T^{'}(z)\neq 0$. The set of all $w=T(z)$ with $T^{'}(z)=0$ is the image under $T$ of a discrete set in $\mathscr{P}^2-\mathscr{V}$, hence discrete in $\mathscr{M}^2-T(\mathscr{V})$. Then, by the Riemann removable singularity theorem, $T^{-1}$ is analytic on $\mathscr{P}^2-\mathscr{V}$. Hence, $T$ is analytically diffeomorphic on $\mathscr{P}^2-\mathscr{V}$.$\hfill\qedsymbol$

Since the Jacobian of $T$, $J_T$, is not well defined on $\mathscr{V}$, the boundedness of $\det (J_T)$ guaranteed by the compact set ${\mathscr{P}^2}$ and the continuous map $J_T$ needs to be re-verified. The next proposition shows that $\det (J_T)$ is bounded on $\mathscr{P}^2-\mathscr{V}$.

\textit{Proposition 4:} The Jacobian determinant of $T:\mathscr{P}^2\to\mathscr{M}^2$ is bounded on $\mathscr{P}^2-\mathscr{V}$.

\textit{Proof:} %See the proof of Proposition 4 in \cite{FanLi2022concers}.
From the boundary behavior of $T$, we know that the boundedness of $T$ on $\mathscr{P}^2-\mathscr{V}$ is the same as that on $\mathscr{O}_0-\mathscr{V}_0$\cite{pommerenke1992}. Hence, it is sufficient to show that $T$ is bounded on $\mathscr{O}_0-\mathscr{V}_0$. Let the consecutive vertices be $z_1$,...,$z_n$ in the counterclockwise direction. The angle at $z_k$ is given by the value of $\arg\frac{z_{k-1}-z_k}{z_{k+1}-z_k}$, denote by $\alpha_k\pi, 0<\alpha_k<2.$ Let the circular sector $S_k$ be the intersection of a sufficiently small disk around $z_k$ with $\mathscr{P}^2$. For $S_k$, a single-valued branch of $\phi_{z_k}(\zeta)=(z-z_k)^{1/\alpha_k}$ maps $S_k$ onto a half disk $U_k^+$. A suitable branch of $\phi_{z_k}^{-1}=z_k+\zeta^{\alpha_k}$ maps $U_k^+$ onto $S_k$, and we may consider the function $G(\zeta)=T(z_k+\zeta^{\alpha_k})$ in $U_k^+$. According to the Reflection Principle \cite{pommerenke1992}, $G(\zeta)$ has an analytic continuation to the whole disk. Thus, $G(\zeta)$ is analytic at the origin and its Taylor expansion is given by $G(\zeta)=T(z_k+\zeta^{\alpha_k})=T(z_k)+\sum_{m=1}^\infty a_m\zeta^m, a_1\neq 0$. Inverting the series in a neighborhood of $T(z_k)$ and on setting $w=T(z_k+\zeta^{\alpha_k})$, we have $\zeta=\sum_{m=1}^\infty b_m (w-w_k)^m, b_1\neq 0$, where $w_k=e^{i\theta_k}$ is the image of vertex $z_k$ under $T$ in the unit circle $C$. Taking exponentials, we obtain $T^{-1}(w)-z_k=(w-w_k)^{\alpha_k}E_k(w)$, where $E_k(w)$ is analytic and non-zero near $w_k$. Taking the derivative, we conclude that the product $F(w)=(T^{-1})^{'}(w)\prod_{k=1}^{n}(w-w_k)^{1-\alpha_k}$ is analytic and non-vanishing in the closed unit disk $\overline{D}$. 

Next consider the boundedness of $|(T^{-1})^{'}(w)|$ on $\overline{D}-T(\mathscr{V}_0)$. %Next consider the argument of $F(w)$, $\arg F(w)$, on the circular arc from $w_k$ to $w_{k+1}$. 
Since $T$ is analytic on $\mathscr{P}^2-\mathscr{V}$ and homeomorphic on $\partial\mathscr{P}^2$, $T^{-1}$ smoothly maps the anticlockwise arc from $w_k$ to $w_{k+1}$ to the line segment from $z_k$ to $z_{k+1}$, with derivative non-vanishing. Thus on taking arguments $$\arg\frac{d}{d\theta}(T^{-1})(e^{i\theta})=\arg(z_{k+1}-z_{k})$$ and thus by the chain rule $$\arg (T^{-1})^{'}(e^{i\theta})=\arg(z_{k+1}-z_{k})-(\theta+\frac{\pi}{2}).$$ For the factor $w-w_k$, we have that $$w-w_k=2\sin \frac{1}{2}(\theta-\theta_k)e^{i(\theta+\theta_k+\frac{\pi}{2})},$$ which implies that its argument is $\theta$ plus a constant. Thus, we conclude that $$\arg\prod_{k=1}^{n}(w-w_k)^{1-\alpha_k}=\frac{1}{2}\theta\cdot\sum_1^n (1-\alpha_k)+C_1,$$ where $C_1$ is a constant. From the fact that $\pi\cdot\sum_1^n (1-\alpha_k)$ is the sum of the exterior angles of a polygon, we have that $$\sum_1^n (1-\alpha_k)=2.$$ and $$\arg F(w)=\arg(z_{k+1}-z_{k})-(\theta+\frac{\pi}{2})+\theta+C_1=C_2,$$ where $C_2=\arg(z_{k+1}-z_{k})-\frac{\pi}{2}+C_1$ is a constant. Hence, $\arg F(w)$ is constant between $w_k$ and $w_{k+1}$. Then, the function $\arg{F(w)}=\Im \log F(w)$ is harmonic on $D$, continuous on $\overline{D}$, and constant on $C$. According to the maximum principle, we conclude that $\arg F(w)=\Im\log F(w)$ is constant on $D$, and from the Cauchy-Riemann equations, $F(w)$ is also constant. Hence, for $T^{-1}$ on $\overline{D}-T(\mathscr{V})$ , we have that $$(T^{-1})^{'}(w)=C_3\prod_{k=1}^{n}(w-w_k)^{\alpha_k-1},$$ where $C_3$ is a constant. Since $w\neq w_k$ for $w\in\overline{D}-T(\mathscr{V}_0)$, $|(T^{-1})^{'}(w)|$ is bounded.

Last consider the boundedness of $\det(J_T(x))$ on $\mathscr{O}_0-\mathscr{V}_0$. Since $|(T^{-1})^{'}(w)|$ is bounded on $\overline{D}-T(\mathscr{V}_0)$, then according to inverse mapping theorem, $|T^{'}(z)|$ is bounded on $\mathscr{O}_0-\mathscr{V}_0$. By the Cauchy-Riemann equations, we have that $$\det(J_T(x))=|T^{'}(z)|^2.$$ Finally, since $|T^{'}(z)|$ is bounded on $\mathscr{O}_0-\mathscr{V}$, the Jacobian determinant of $T$ is bounded on $\mathscr{O}_0-\mathscr{V}_0$.$\hfill\qedsymbol$

%We have thus far established the analyticity of $T$ and the boundedness of $\det (J_T)$ on $\mathscr{P}^2-\mathscr{V}$. The final theorem shows that the conformal navigation transformation between $\mathscr{P}^2$ and $\mathscr{M}^2$ does not change the navigation function properties.

\section{Controller Design Using The Conformal Navigation Transformation}

In this section, a provably correct feedback controller based on the conformal navigation transformation is presented and analyzed to illustrate the application of the proposed transformation to the robot navigation problem in complex 2D polygonal environments.

The conformal navigation transformation provides a geometric approach to transforming two navigation problems with different geometric details into the same problem, as described in the following theorem.
%The invariance of navigation properties is important for extending the construction of navigation functions on sphere worlds to any complex polygonal 2D environment, as described in theorem 1.

\textit{Theorem 1 :} Let $\mathscr{P}^2\subset\mathbb{R}^2$ be a polygonal workspace, and $V\subset\partial\mathscr{P}^2$ be the set of vertices. Let $\phi:\mathscr{M}^2\to[0,1]$ be a navigation function on $\mathscr{M}^2$.  If $T:\mathscr{P}^2\to\mathscr{M}^2$ is the unique conformal navigation transformation determined by $\mathscr{P}^2$, then $\tilde{\phi}=\phi\circ T$ is a navigation function on $\mathscr{P}^2$.

\textit{Proof:} The continuity of $\tilde{\phi}$ on $\mathscr{P}^2$ follows from the fact that both $\phi$ and $T$ have this property. From Lemma 1, we know that $T$ is analytic on $\mathscr{P}^2-\mathscr{V}$. Then, $\tilde{\phi}$ is an analytic function on $\mathscr{P}^2-\mathscr{V}$ because it is a composition of two analytic functions.

Since $T$ is an analytic diffeomorphism on $\mathscr{P}^2-\mathscr{V}$, the Jacobian of $T$ is non-singular in $\mathscr{P}^2-\mathscr{V}$.  Applying the chain rule yields $\nabla\tilde{\phi}=[J_T]^T\nabla(\phi\circ T)$. 
According to Proposition 4, the gradient $\nabla\tilde{\phi}$ is bounded on $\mathscr{P}^2-\mathscr{V}$.

As $J_T$ is non-singular in $\mathscr{P}^2-\mathscr{V}$, $T$ is a bijection from the set of critical points of $\tilde{\phi}$, $\mathscr{C}_{\tilde{\phi}}$, to the set of critical points of $\phi$, $\mathscr{C}_{\phi}$. Furthermore, the Hessian matrix of $\tilde{\phi}$ at the critical points satisfies $H(\tilde{\phi})=[J_T]^T H(\phi\circ T)[J_T]$. Since $T$ is non-singular in $\mathscr{P}^2-\mathscr{V}$, $H(\tilde{\phi})$ is non-singular at the critical points. Hence, $\tilde{\phi}$ is a Morse function on $\mathscr{P}^2-\mathscr{V}$.

Next, consider the image of $\partial\mathscr{O}_i$ under $\tilde{\phi}$. Since $T$ is an analytic diffeomorphism on $\mathscr{P}^2-\mathscr{V}$ and a homeomorphism on each polygonal boundary $\partial\mathscr{O}_i$, the image of $\partial\mathscr{O}_i$ under $T$ is exactly $\partial\mathscr{\tilde{O}}_i$. Moreover, $\phi$ is uniformly maximal on $\partial\mathscr{O}_i$. Hence, $\tilde{\phi}$ is admissible on $\mathscr{P}^2$.

Finally, let us next verify the polar
property of $T$ in $\mathscr{P}^2$. At the critical points, we have that: $H(\tilde{\phi})=[J_T]^T H(\phi\circ T)[J_T]$. Since $T$ is a bijection between $\mathscr{C}_{\tilde{\phi}}$ and $\mathscr{C}_{\phi}$, the Morse index of $\tilde{\phi}$ matches the Morse index of $\phi$ at each of the critical points. Moreover, $\phi$ has a unique minimum at $x_d$ in $\mathscr{P}^2-\mathscr{V}$. Hence, $\tilde{\phi}$ is a polar function at $p_d=T(x_d)$ in $\mathscr{P}^2-\mathscr{V}$. Since the admissibility of $T$ on $\mathscr{P}^2$ means that $\tilde{\phi}$ is uniformly maximal on $\partial\mathscr{P}^2$, any vertex in  $\mathscr{V}\subset \partial\mathscr{P}^2$ is not a local minimum. It follows that $\tilde{\phi}$ satisfies the polar property on $\mathscr{P}^2$.$\hfill\qedsymbol$

Theorem 1 shows that the composition of the conformal navigation transformation $T:\mathscr{P}^2\to\mathscr{M}^2$ with a navigation function on $\mathscr{M}^2$ forms a navigation function on $\mathscr{P}^2$. This property provides a method for designing feedback control laws to achieve convergence from almost all initial points to the goal point, as shown below.  

\textit{Theorem 2:} The mobile robot system \eqref{kinematic} under the feedback control law \begin{equation}\begin{split}\label{}u(x)&=-K(J_T(x))^T\nabla_{T(x)}\phi_{kr}(T(x))\\&=-K\lVert J_T(x)\rVert(J_T(x))^{-1}\nabla_{T(x)}\phi_{kr}(T(x)),\end{split}\label{Theorem 2 control law }\end{equation} where $K$ is a positive gain, is globally asymptotically stable at the goal point $x_d\in\mathring{\mathscr{P}^2}$ from almost every initial point in $\mathscr{P}^2-\mathscr{V}$.

\textit{Proof:} Since $T$ is an analytic diffeomorphism on $\mathscr{P}^2-\mathscr{V}$, according to the Cauchy-Riemann equation, the Jacobian $J_T(x)=\begin{bmatrix}{u_x}_1&-{v_x}_1\\{v_x}_1&{u_x}_1\end{bmatrix}$ at $x=(x_1,x_2)\in{\mathscr{P}^2-\mathscr{V}}$. Moreover, the Jacobian $J_T(x)={({{u_x}_1}^2+{{v_x}_1}^2)}^{\frac{1}{2}}R={\lVert J_T(x)\rVert}^{\frac{1}{2}}R$, where $R$ is a rotation matrix. Hence, for $T$ on $\mathscr{P}^2-\mathscr{V}$, we have that: $(J_T(x))^T=\lVert J_T(x)\rVert(J_T(x))^{-1}$. 

Choosing $V(x)=\phi_{kr}(T(x))$ as a candidate Lyapunov function. Differentiating $V$ along the system's trajectories yields: $\dot{V}(x)=-K\lVert J_T(x)\lVert\lVert\nabla_{T(x)}\phi_{kr}(T(x))\lVert^2\leqslant0.$
Since $J_T(x)$ is non-singular, the set for $\dot{V}(x)=0$ consists only of the critical points of $\phi_{kr}(T(x))$. The critical points of $\phi_{kr}(T(x))$ include the unique minimum point $x_d$ and the isolated saddle points. Lasalle's Invariance Theorem dictates that the closed loop system will converge to the largest positive invariant set that consists only of the goal and the saddle points. Since a saddle point has a stable manifold of 1-dimension in $\mathring{\mathscr{P}^2}$, all isolated saddle points have attractive basins of zero measure. Hence, $\dot{V}(x)<0$ holds almost everywhere, which implies that the system is globally asymptotically stable at $x_d$ from almost every initial point in $\mathscr{P}^2-\mathscr{V}$.$\hfill\qedsymbol$

\section{Construction Of The Conformal Navigation Transformation}

In \cite{FanLi2022}, a construction method for the conformal navigation transformation is proposed on a 2-dimensional workspace with smooth boundaries. Here, we will present a construction method for $T$ that operates on a polygonal workspace. The first step is to construct $T$ on a simply connected polygonal workspace. The second step is to use the method proposed in \cite{FanLi2022} to obtain $T$ on a multiply connected polygonal workspace.

Let $U^+$ and $U^-$ be a bounded and an unbounded simply connected workspace in the extended complex plane $\mathbb{C}\cup\{\infty\}$, respectively. 
For $U^+$, assume that $z_c$ is a given point in $U^+$. For $U^-$, assume that $z_c$ is a given point in the complement of $U^-$ and $\infty\in U^-$. The boundary $\Gamma$ is a closed polygonal curve parametrized by a $2\pi$-periodic complex analytic function $\gamma(s)$ except at $n$ points \begin{equation} s_k=(k-1)\frac{2\pi}{n} \in[0,2\pi],\ k\in\{1,2,...,n\}.\label{sk}\end{equation} The orientation of $\Gamma$ is counterclockwise for $U^+$ and clockwise for $U^-$. 

%According to Definition 4, the conformal navigational transformation $T(z)$ in the extended complex plane is a holomorphic function with a non-zero derivative and has a continuous extension to the boundary $\Gamma$. Then, from the Cauchy integral formula we have that $T(z)$ on $U^+$ and $U^-$ are completely determined by their values on boundary $\Gamma$. This property reduces the construction problem of $T(z)$ on the entire workspace to the construction problem of $T(\gamma (s))$ on the boundary. 
The construction method for $T(z)$ on $U^+$ with a smooth boundary is given by the following results. Moreover, the described method can be straightforwardly applied to solve $T(z)$ on $U^-$. For more details, see \cite{FanLi2022}.

For a bounded simply connected workspace $U^+$ with a smooth boundary, $T(z)$ is constructed as follows: \begin{equation} T(z)=\dot{T}(z_c)(z-z_c)e^{(z-z_c)f(z)},\label{T(z)}\end{equation} where $f(z)$ is a holomorphic function on $U^+$.
The boundary values of the function $f(z)$ are given by: \begin{equation}G(s)f(\gamma (s))=\mu(s)+c(s)+i\upsilon(s),\label{integral simple}\end{equation} where $G(s)=\gamma (s)-z_c$, $\mu(s)=-\log|\gamma (s)-z_c|$, $c(s)=-\log\big(\dot{T}(z_c)\big)$, $\upsilon(s)=\theta (s)-\arg\big(\gamma (s)-z_c\big)$. The unknown functions $c$ and $\upsilon$ are uniquely determined by the known $\mu$. The following proposition describes the above conclusion in more detail. The various variables mentioned in the sequel are defined in \cite{FanLi2022}.

\textit{Proposition 5 \cite{FanLi2022}:} For the given real-valued function $\mu$, there exists a unique constant $c$ and a unique function $\upsilon$, such that \eqref{integral simple} are boundary values of a holomorphic function $f$ in $U^+$. The function $\upsilon$ is the unique solution of the equation \begin{equation} \mu-\mathbf R\mu=-\mathbf H \upsilon,\label{integral equation}\end{equation} and $c$ is given by \begin{equation} c=[\mathbf H \mu-(\mathbf I-\mathbf R)\upsilon]/2.\label{c}\end{equation}

Since \eqref{integral simple} is uniquely solvable, the conformal navigation transformation $T(z)$ on $U^+$ can be solved by \eqref{T(z)}, as described in Proposition 6.

\begin{figure}[!t]\centering
	\includegraphics[width=8cm]{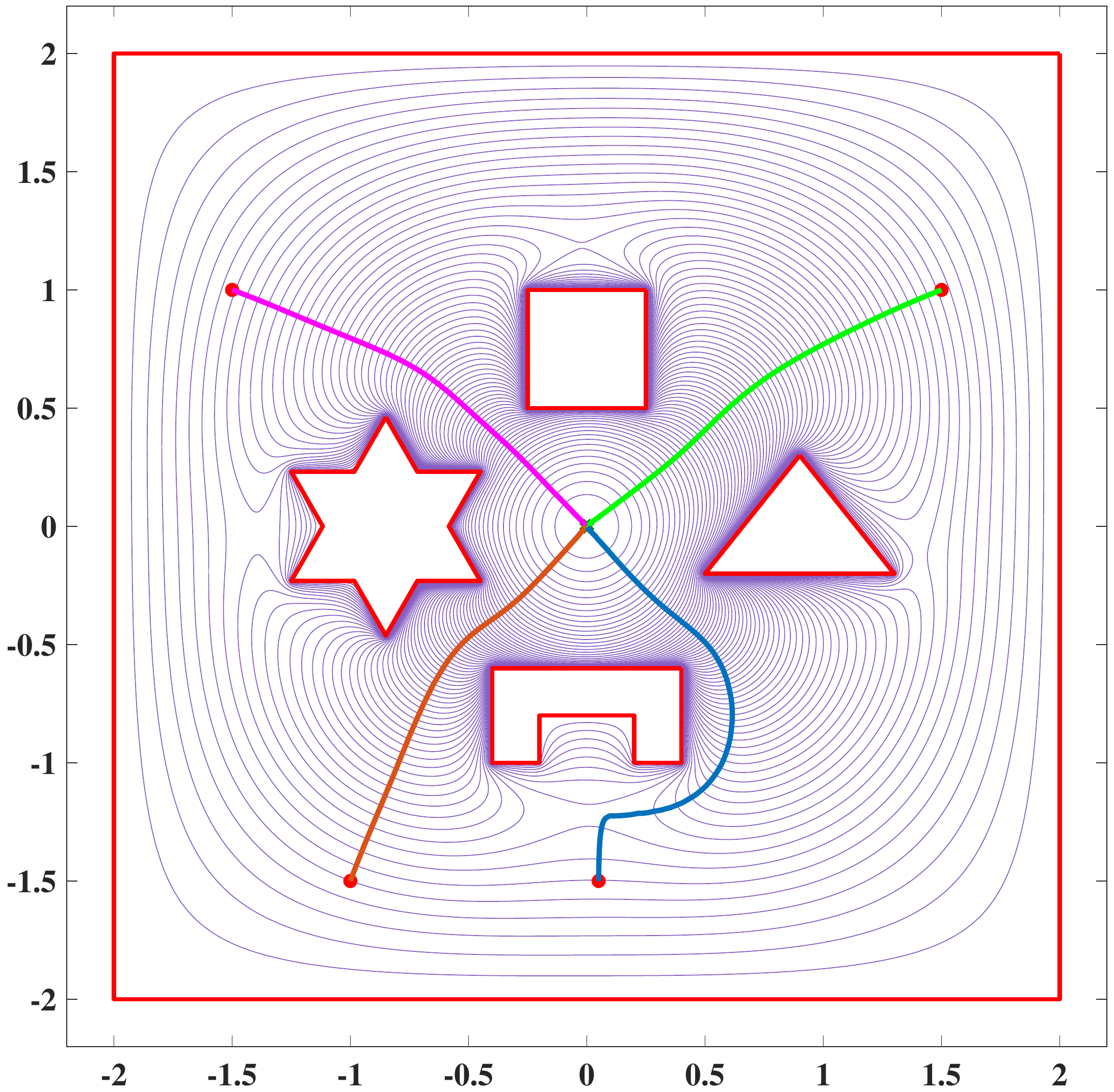}
	\caption{Trajectories of the kinematic mobile robot \eqref{kinematic} under the control law \eqref{Theorem 2 control law } for different initial points in a 2-dimensional polygonal workspace}
	%Trajectory of kinematic robot \eqref{kinematic} under control law \eqref{Proposition 4 control law } in a 2D complex workspace}
	\label{Fig_2}
\end{figure}

\textit{Proposition 6 \cite{FanLi2022}:} For any bounded simply connected workspace $U^+$, and any twice continuously differentiable parametric boundary $\gamma (s)$ with $\dot{\gamma }(s)\neq 0$, the holomorphic function
\begin{equation} T(z)=e^{-c}(z-z_c)e^{(z-z_c)f(z)}\end{equation} is the conformal navigation transformation on $U^+$.

For a bounded simply connected workspace $U^+$ with a polygonal boundary, the curve parametric equation $\gamma(s)$ is nonsmooth at $s_k,\ k\in\{1,2,...,n\}$. Therefore, before using the above results, it is necessary to rewrite equation \eqref{integral equation} and eliminate the singularity at $s_k$.

Since the constant is an eigenfunction of $\mathbf R$ corresponding to -1 and an eigenfunction of $\mathbf H$ corresponding to 0, the equation \eqref{integral equation} can be rewritten as \begin{equation} 2\mu-\mathbf R(s,t)[\mu(t)-\mu(s)]=-\mathbf H(s,t) [\upsilon(t)-\upsilon(s)],\label{integral equation vertex}\end{equation} where $t,s\in[0,2\pi],\ t\neq s$.

Now, define the function \begin{equation} \sigma (s)=2\pi\frac{[g(s)]^q}{[g(s)]^q+[g(2\pi-s)]^q},\label{Strictly monotonically increasing bijective smooth function}\end{equation} where \begin{equation} g(s)=\Big(\frac{1}{q}-\frac{1}{2}\Big)\Big(\frac{\pi-s}{\pi}\Big)^3+\frac{s-\pi}{q\pi}+\frac{1}{2}.\label{g}\end{equation}
The function $\sigma(s)$ is a smooth, strictly monotonically increasing and bijective function. The parameter $q$ is an integer and $p\geq2$.

Define the function \begin{equation} \eta (s)=\frac{1}{n}\sigma\big(n(s-s_k)\big)+s_k,\  s\in[s_k,s_{k+1}].\label{eta}\end{equation}
For the function $\eta(s)$, we have that
$\dot{\eta}(s_k)=0$ and $\dot{\eta}(s)\neq0$ for $s\neq s_k$. 

Finally, define \begin{equation} \hat{\gamma}(\tau)=\gamma\big(\eta(\tau)\big),\  \tau\in[0,2\pi].\label{new gamma}\end{equation}
and consequently we obtain that \begin{equation} \int_0^{2\pi}\hat{\gamma}(\tau)d\tau=\int_0^{2\pi}\gamma(s)ds.\label{int new gamma}\end{equation} Then, the function $\hat{\gamma}$ is smooth on $\Gamma$ and $\hat{\gamma}(0)=\hat{\gamma}(2\pi)=0$. Hence, the singularity at $s_k$ is eliminated. After rewriting Equation \eqref{integral equation} and eliminating the singularity at $s_k$, the solution of $T$ on $U^+$ can be obtained using the construction method of $T$ on $U^+$ with a smooth boundary.

The construction method of $T(z)$ on a multiply connected workspace proposed in \cite{FanLi2022} is an efficient and fast iterative method, which is inspired by the conventional Koebe's method on unbounded regions. Meanwhile, having obtained $T$ on $U^+$ and $U^-$, the method can be applied directly to solving the conformal transformation on a multiply connected polygonal workspace. See section \uppercase\expandafter{\romannumeral5} in \cite{FanLi2022} for details.

\begin{figure}[!t]\centering
	\includegraphics[width=8cm]{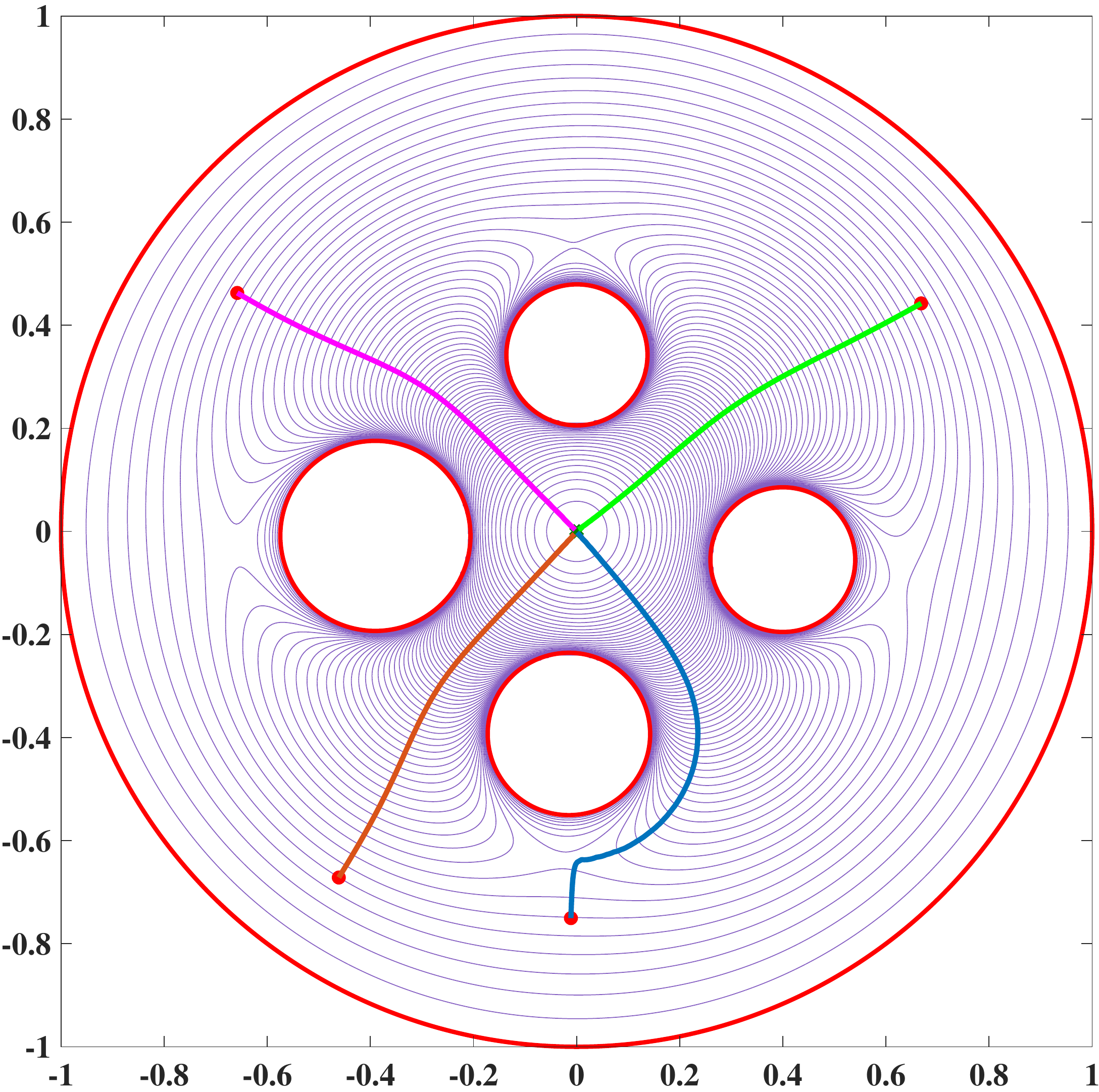}
	\caption{Trajectories of the kinematic mobile robot \eqref{kinematic} under the control law \eqref{Theorem 2 control law } for different initial points in the conformal sphere world.}
	%Trajectory of kinematic robot (4) under control law (6) in the conformal sphere world corresponding to Fig. 2}
	\label{Fig_3}
\end{figure}

\section{SIMULATION RESULTS}

In order to validate the effectiveness of the proposed methodology, a numerical simulation was presented in this section. The simulation was carried out for controller \eqref{Theorem 2 control law }, and the Koditschek-Rimon navigation function $\phi_{kr}$ was used in the sphere world. The parameter $k$ in $\phi_{kr}$ was set as $k=6$. The gain $K$ in \eqref{Theorem 2 control law } was set as $K=1$. For the conformal navigation transformation, we iterated until $\|\hat{T}_n-\hat{T}_{n-1}\|<1\times 10^{-12}$. 

In the simulation, the feedback control law \eqref{Theorem 2 control law } provided in Theorem 2 is applied to the kinematic mobile robot \eqref{kinematic}. The workspace is set up with three star-shaped internal polygonal obstacles: a triangle, a rectangle, a hexagonal star dodecagon; and a non-star-shaped polygonal obstacle that cannot be handled well by the homeomorphic-based transformations. The goal point is set as
$x_d=(0,0)$, which the robot aims to converge to from 4 different
initial points, namely $x_0=(1.5,1),(-1.5,1),(-1,-1.5)$ and $(0.05,-1.5)$. Fig. 1 and Fig. 2 depict the level sets of the navigation function and robot trajectories in the initial workspace and the transformed sphere world, respectively.
%In Figs. 2 and Figs. 3, we observe the level sets of the navigation function and robot trajectories in the initial workspace and the transformed sphere world, respectively. 
As we can observe, the designed feedback control law successfully performs the task of maintaining the robot in the initial workspace and the transformed sphere world, avoiding collisions and stabilizing it at the goal point. In this example, the number of iterations is 7 and $\|\hat{T}_{7}-\hat{T}_{6}\|=1.5831\times 10^{-13}$.

\section{CONCLUSION}

This paper proposes a methodology for the navigation problem in complex polygonal workspaces based on a novel spatial transformation called the conformal navigation transformation. The class of navigation functions defined in smooth workspaces is extended to accommodate polygonal obstacles. With the application of the proposed transformation, the navigation function in the geometrically simple sphere world is pulled back to the initial workspace. The negative gradient of the pulled-back navigation function then serves as a control law that enables the dynamic robotic system to converge from almost all initial points to the goal point. It is shown that this method provides solutions for both path planning and motion planning subproblems. Furthermore, an iterative method for constructing the proposed transformation is presented, which ensures fast convergence and is suitable for non-star-shaped polygonal obstacles. Finally, simulation results support the theoretical results.

Future research directions include extending the solution to account for dynamic obstacles, multi-agent navigation problems, and more realistic scenarios with local sensing.

\bibliographystyle{IEEEtran}
\bibliography{Bibliography/IEEEabrv,Bibliography/Mybib}\ %IEEEabrv instead of IEEEfull

\begin{thebibliography}{10}
\providecommand{\url}[1]{#1}
\csname url@samestyle\endcsname
\providecommand{\newblock}{\relax}
\providecommand{\bibinfo}[2]{#2}
\providecommand{\BIBentrySTDinterwordspacing}{\spaceskip=0pt\relax}
\providecommand{\BIBentryALTinterwordstretchfactor}{4}
\providecommand{\BIBentryALTinterwordspacing}{\spaceskip=\fontdimen2\font plus
\BIBentryALTinterwordstretchfactor\fontdimen3\font minus
  \fontdimen4\font\relax}
\providecommand{\BIBforeignlanguage}[2]{{%
\expandafter\ifx\csname l@#1\endcsname\relax
\typeout{** WARNING: IEEEtran.bst: No hyphenation pattern has been}%
\typeout{** loaded for the language `#1'. Using the pattern for}%
\typeout{** the default language instead.}%
\else
\language=\csname l@#1\endcsname
\fi
#2}}
\providecommand{\BIBdecl}{\relax}
\BIBdecl

\bibitem{1}
H.~Choset \emph{et~al.}, \emph{Principles of robot motion}.\hskip 1em plus
  0.5em minus 0.4em\relax Cambridge, MA, USA: MIT press, 2005.

\bibitem{2}
O.~Khatib, ``Real-time obstacle avoidance for manipulators and mobile robots,''
  in \emph{Proc. IEEE Int. Conf. Robot. Autom.}, 1985, pp. 500--505.

\bibitem{Rimon1992}
E.~Rimon and D.~E. Koditschek, ``Exact robot navigation using artificial
  potential functions,'' \emph{IEEE Trans. Robot. Autom.}, vol.~8, no.~5, pp.
  501--518, Oct 1992.

\bibitem{6}
------, ``The construction of analytic diffeomorphisms for exact robot
  navigation on star worlds,'' \emph{Trans. Am. Math. Soc.}, vol. 327, no.~1,
  pp. 71--116, 1991.

\bibitem{7}
D.~E. Koditschek and E.~Rimon, ``Robot navigation functions on manifolds with
  boundary,'' \emph{Adv. Appl. Math.}, vol.~11, no.~4, pp. 412--442, 1990.

\bibitem{Loizou2008}
S.~G. Loizou and K.~J. Kyriakopoulos, ``Navigation of multiple kinematically
  constrained robots,'' \emph{IEEE Trans. Robot.}, vol.~24, no.~1, pp.
  221--231, feb 2008.

\bibitem{11}
H.~G. Tanner and A.~Boddu, ``{Multiagent navigation functions revisited},''
  \emph{IEEE Trans. Robot.}, vol.~28, no.~6, pp. 1346--1359, Dec 2012.

\bibitem{12}
C.~K. Verginis, Z.~Xu, and D.~V. Dimarogonas, ``Decentralized motion planning
  with collision avoidance for a team of uavs under high level goals,'' in
  \emph{Proc. IEEE Int. Conf. Robot. Autom.}, 2017, pp. 781--787.

\bibitem{Li2019}
C.~Li and H.~G. Tanner, ``Navigation functions with time-varying destination
  manifolds in star worlds,'' \emph{IEEE Trans. Robot.}, vol.~35, no.~1, pp.
  35--48, feb 2019.

\bibitem{Loizou2022}
S.~G. Loizou and E.~D. Rimon, ``Mobile robot navigation functions tuned by
  sensor readings in partially known environments,'' \emph{IEEE Robot. Autom.
  Lett.}, vol.~7, no.~2, pp. 3803--3810, 2022.

\bibitem{15}
C.~K. Verginis and D.~V. Dimarogonas, ``Adaptive robot navigation with
  collision avoidance subject to 2nd-order uncertain dynamics,''
  \emph{Automatica}, vol. 123, p. 109303, jan 2021.

\bibitem{16}
S.~Berkane and D.~V. Dimarogonas, ``Constrained stabilization on the
  n-sphere,'' \emph{Automatica}, vol. 125, p. 109416, mar 2021.

\bibitem{17}
S.~Paternain, D.~E. Koditschek, and A.~Ribeiro, ``Navigation functions for
  convex potentials in a space with convex obstacles,'' \emph{IEEE Trans.
  Automat. Contr.}, vol.~63, no.~9, pp. 2944--2959, sep 2018.

\bibitem{18}
P.~Vlantis, C.~Vrohidis, C.~P. Bechlioulis, and K.~J. Kyriakopoulos, ``Robot
  navigation in complex workspaces using harmonic maps,'' in \emph{Proc. IEEE
  Int. Conf. Robot. Autom.}, 2018, pp. 1726--1731.

\bibitem{19}
S.~G. Loizou, ``The navigation transformation,'' \emph{IEEE Trans. Robot.},
  vol.~33, no.~6, pp. 1516--1523, dec 2017.

\bibitem{Nicolas2020}
N.~Constantinou and S.~G. Loizou, ``Robot navigation on star worlds using a
  single-step navigation transformation,'' in \emph{Proc. IEEE Conf. Decis.
  Control}, 2020, pp. 1537--1542.

\bibitem{20}
D.~E. Blair, \emph{Inversion theory and conformal mapping}.\hskip 1em plus
  0.5em minus 0.4em\relax Providence, RI, USA: Amer. Math. Soc., 2000.

\bibitem{FanLi2022}
L.~Fan, J.~Liu, W.~Zhang, and P.~Xu, ``Conformal navigation transformations
  with application to robot navigation in complex workspaces,'' \emph{arXiv
  preprint arXiv:2208.06876}, 2022.

\bibitem{pommerenke1992}
C.~Pommerenke, \emph{Boundary behaviour of conformal maps}.\hskip 1em plus
  0.5em minus 0.4em\relax Berlin, Germany: Springer-Verla, 1992, vol. 299.

\end{thebibliography}

\vspace{12pt}
\color{red}

\end{document}